%% file: lstm-pynq.tex
\documentclass[a4paper,conference]{ieee/IEEEtran}

\usepackage{amssymb}
\usepackage[cmex10]{amsmath}

\usepackage[capitalize]{cleveref}

\usepackage{float}

\usepackage{graphicx}

\usepackage{subcaption}

\usepackage{calc}

\usepackage{wrapfig}

\usepackage{tabularx}
\usepackage{booktabs}
\usepackage{footnote}

\usepackage{makeidx}        % allows index generation

\usepackage{balance} % use \balance command in left column of the last page.

\usepackage{textcomp}

\usepackage{booktabs}

\usepackage{color}

\usepackage[acronym]{glossaries}
\newglossary[symlog]{symbol}{symi}{symo}{Symbols}
\glstoctrue 
\loadglsentries[\acronymtype]{tex/acronyms}

\usepackage{mathtools}
\usepackage{amsmath}

\usepackage{threeparttable}

\renewcommand{\baselinestretch}{0.975}

\begin{document}

% \title{Analysis of Design Trade-offs for Variable Precision LSTM Networks on FPGAs}
\title{FINN-L: Library Extensions and Design Trade-off Analysis for Variable Precision LSTM Networks on FPGAs}
% \title{Exploration of Accuracy, Hardware Cost and Throughput vs. Precision of Bidirectional LSTM Network on FPGA}
% \title{An Open Source Library of Fully Paremetrisable LSTM for FPGA }

\author{
\IEEEauthorblockN{Vladimir Rybalkin, \\Muhammad Mohsin Ghaffar and Norbert Wehn}
\IEEEauthorblockA{Microelectronic Systems Design Research Group\\
University of Kaiserslautern, Germany\\
\{rybalkin, ghaffar, wehn\}@eit.uni-kl.de}
\vspace{-5ex}
\and
\IEEEauthorblockN{Alessandro Pappalardo, \\Giulio Gambardella and Michaela Blott}
\IEEEauthorblockA{Xilinx Research Labs\\
\{alessand, giuliog, mblott\}@xilinx.com}
\vspace{-5ex}}
    
%\newcommand{\TODO}[1]{\todo[inline]{#1}}
%\author{\\ \\ \\ \\--blind review--\\ \\ \\ \\}

%\copyrightnotice
\input{tex/copyrightnotice}
\maketitle

\input{tex/abstract}

\input{tex/introduction}

\input{tex/related_works}

\input{tex/theory}
\input{tex/training}
\input{tex/inference}

\input{tex/results}

\input{tex/conclusion}

\section*{acknowledgment}
This work was partially supported by the InnoProm program of the state Rhineland-Palatinate, Germany. We also thank Insiders Technologies GmbH for consultancy in machine learning.

%\vspace{-1.5em} 

\bibliographystyle{ieee/IEEEtran} 
\bibliography{ieee/IEEEexample}

\end{document}

%% file: tex/acronyms.tex
\newcommand{\newacronymf}[3]{\newglossaryentry{#1}{name={#2},description={#3},first={#2}}}

%================= 0123 ==================
\newacronymf{3gpp}{3GPP}{Third Generation Partnership Project}
\newacronymf{4g}{4G}{fourth generation of mobile phone system specifications}
\newacronymf{5g}{5G}{fifth generation of mobile phone system specifications}

%================= A ==================
\newacronym{ann}{ANN}{Artificial Neural Network}
\newacronym{anns}{ANNs}{Artificial Neural Networks}
\newacronym{acq}{ACQ}{acquisition}
\newacronym{app}{APP}{a posteriori probability}
\newacronym{ask}{ASK}{amplitude-shift keying}
\newacronym{awgn}{AWGN}{Additive White Gaussian Noise}
\newacronym{arq}{ARQ}{automatic repeat request}
\newacronym{acs}{ACS}{add-compare-select}
\newacronym{asic}{ASIC}{application specific integrated circuit}
\newacronym{arp}{ARP}{almost regular permuatation}
\newacronym{asip}{ASIP}{application specific instruction set processor}
\newacronymf{amba}{AMBA}{Advanced Microcontroller Bus Architecture}
\newacronym{ahb}{AHB}{advanced high-performance bus}
\newacronym{aacs}{AACS}{a priori aware constant sphere}
\newacronym{ant}{ANT}{algorithmic noise-tolerance}

%================= B ==================
\newacronym{bhl}{BHL}{Backward Hidden Layer}
\newacronym{blstm}{BLSTM}{Bidirectional Long Short-Term Memory}
\newacronym{bcjr}{BCJR}{Bahl, Cocke, Jelinek, and Raviv}
\newacronym{bpsk}{BPSK}{Binary Phase Shift Keying}
\newacronym{ber}{BER}{bit error rate}
\newacronym{bp}{BP}{Belief Propagation}
\newacronym{bw}{BW}{Bit Width}
\newacronym{bicm}{BICM}{bit interleaved coded modulation}
\newacronym{blast}{BLAST}{Bell Labs layered space time}
\newacronymf{bc}{BC}{best case, 1.3V, -40$^\circ$C}
\newacronym{bilstm}{BiLSTM}{Bidirectional LSTM}

%================= C ==================
\newacronym{ctc}{CTC}{Connectionist Temporal Classification}
\newacronym{cn}{CN}{Check Node}
\newacronym{cr}{CR}{Code Rate}
\newacronym{cfu}{CFU}{Check Node Functional Unit}
\newacronym{crc}{CRC}{Cyclic Redundancy Check}
\newacronymf{cdma2000}{CDMA2000}{name of a communication standard}
\newacronym{cc}{CC}{convolutional code}
\newacronym{cnb}{CNB}{check node block}
\newacronym{cnp}{CNP}{check node processor}
\newacronym{cdma}{CDMA}{code division multiple access}
\newacronymf{cmos}{CMOS}{complementary metal oxide semiconductor}
\newacronymf{cpu}{CPU}{central processing unit}
\newacronymf{cnn}{CNN}{Convolutional Neural Network}
\newacronymf{cnns}{CNNs}{Convolutional Neural Networks}

%================= D ==================
\newacronym{dma}{DMA}{Direct Memory Access}
\newacronym{db}{dB}{decibel}
\newacronym{dp}{DP}{dual ported (memory)}
\newacronym{dram}{DRAM}{dynamic random access memory}
\newacronym{dc}{$d_c$}{CN degree}
\newacronym{dv}{$d_v$}{VN degree}
\newacronym{dvs}{DVS}{dynamic voltage scaling}
\newacronym{dvb}{DVB}{digital video broadcast}
\newacronymf{dvbrcs}{DVB-RCS}{digital video broadcasting - return channel over satellite}
\newacronymf{dvbrct}{DVB-RCT}{digital video broadcasting - return channel terrestrial}
\newacronymf{dvbc2}{DVB-C2}{digital video broadcasting - cable - second generation \cite{eudigital}}
\newacronymf{dvbs2}{DVB-S2}{digital video broadcasting - satellite - second generation \cite{eudigital}}
\newacronymf{dvbs2x}{DVB-S2X}{digital video broadcasting - satellite - extension of the second generation \cite{eudigital}}
\newacronymf{dvbt2}{DVB-T2}{digital video broadcasting - terrestrial - second generation \cite{eudigital}}
\newacronym{dvfs}{DVFS}{dynamic voltage and frequency scaling}
\newacronym{drp}{DRP}{dithered relative prime}
\newacronymf{dfg}{DFG}{Deutsche Forschungsgesellschaft}

%================= E ==================
\newacronym{esf}{ESF}{extrinsic scaling factor}
\newacronym{edc}{EDC}{error detection and correction}
\newacronym{edge}{EDGE}{name of a communications standard}
\newacronym{eds}{EDS}{error detection sequential}
\newacronym{ecc}{ECC}{error correction code}
\newacronym{exit}{EXIT}{extrinsic information transfer}
\newacronym{ems}{EMS}{Extended Min-Sum}
\newacronym{ecn}{ECN}{Elementary Check Node}

%================= F ==================
\newacronym{fhl}{FHL}{Forward Hidden Layer}
\newacronym{fsm}{FSM}{finite state machine}
\newacronym{fec}{FEC}{forward error correction}
\newacronym{fer}{FER}{Frame Error Rate}
\newacronymf{fifo}{FIFO}{first in first out memory}
\newacronymf{fpga}{FPGA}{field-programmable gate array}
\newacronym{fsd}{FSD}{Fixed complexity sphere decoder}
\newacronym{fwbw}{FWBW}{Forward-Backward}

%================= G ==================
\newacronymf{gsm}{GSM}{Global System for Mobile Communication; communication standard}
\newacronym{gpp}{GPP}{general-purpose processor}
\newacronym{gprs}{GPRS}{name of a communications standard}
\newacronym{gps}{GPS}{global positioning system}
\newacronymf{geq}{GE}{gate equivalent}
\newacronym{gfq}{GF$(q)$}{Galois Field}

%================= H ==================
\newacronymf{hspa}{HSPA}{high speed packet access}
\newacronym{harq}{HARQ}{hybrid automatic repeat request}
\newacronym{hw}{HW}{hardware}

%================= I ==================
\newacronym{ic}{IC}{integrated circuit}
\newacronym{io}{I/O}{input / output}
\newacronym{ip}{IP}{intellectual property}
\newacronym{ier}{IER}{injected error rate}
\newacronym{itrs}{ITRS}{International Technology Roadmap for Semiconductors}

%================= J ==================

%================= K ==================

%================= L ==================
\newacronym{lstm}{LSTM}{Long Short-Term Memory}
\newacronym{llr}{LLR}{Logarithmic Likelihood Ratio}
\newacronym{ldr}{LDR}{Logarithmic Density Ratio}
\newacronymf{lte}{LTE}{long term evolution of the 3GPP standard}
\newacronym{ldpc}{LDPC}{Low-Density Parity-Check}
\newacronymf{logmap}{Log-MAP}{MAP algorithm in the logarithmic domain}
\newacronym{lan}{LAN}{local area network }
\newacronym{lsb}{LSB}{least significant bit}
\newacronym{lut}{LUT}{look-up table}
\newacronym{lfsr}{LFSR}{linear feedback shift register}
\newacronym{lord}{LORD}{layered orthogonal lattice detector}
\newacronym{lsd}{LSD}{List sphere decoder}
\newacronym{ldmc}{LDMC}{low-density MIMO check}

%================= M ==================
\newacronym{mac}{MAC}{Multiply-Accumulate}
\newacronym{map}{MAP}{maximum a posteriori probability}
\newacronymf{mlm}{Max-Log-MAP}{MAP algorithm in the logarithmic domain with simplified computation}
\newacronym{mimo}{MIMO}{multiple input multiple output}
\newacronym{mtbf}{MTBF}{mean time between failure}
\newacronym{msb}{MSB}{most significant bit}
\newglossaryentry{mpsoc}{name=MPSoC, description=multi-processor system-on-chip, first=multi-processor system-on-chip (MPSoC), firstplural=multi-processor systems-on-chip (MPSoC)}
\newacronym{mops}{MOPS}{million operations per second}
\newacronym{mmse}{MMSE}{minimum mean square error}
\newacronym{mcmc}{MCMC}{Markov Chain Monte Carlo}
\newacronym{ml}{ML}{maximum likelihood}

%================= N ==================
\newacronym{nns}{NNs}{Neural Networks}
\newacronym{nn}{NN}{Neural Network}
\newacronym{nbldpc}{NB-LDPC}{Non-Binary Low-Density Parity-Check}
\newacronym{nii}{NII}{next iteration initialization}
\newacronym{nsc}{NSC}{non-systematic and non-recursive convolutional}
\newacronym{noc}{NoC}{network on chip}
\newacronymf{nom}{NOM}{nominal case, 1.2V, 25$^\circ$C}

%================= O ==================
\newacronym{ocr}{OCR}{Optical Character Recognition}
\newacronym{otx}{OTX}{outer receiver}
\newacronym{ofdm}{OFDM}{orthogonal frequency division multiplexing}

%================= P ==================
\newacronym{pl}{PL}{Programmable Logic}
\newacronym{pc}{PS}{Processing System}
\newacronym{psk}{PSK}{phase-shift keying}
\newacronym{psmap}{PSMAP}{parallel SMAP}
\newacronym{pr}{P\&R}{placement and routing}
\newacronym{ped}{PED}{partial Euclidean distance}
\newacronym{pic}{PIC}{parallel interference cancellation}
\newacronym{pn}{PN}{Permutation Node}

%================= Q ==================
\newacronym{qam}{QAM}{quadrature amplitude modulation}
\newacronym{qpsk}{QPSK}{quadrature phase shift keying}
\newacronym{qpp}{QPP}{quadratic permutation polynomial}
\newacronym{qos}{QoS}{Quality of Service}

%================= R ==================
\newacronym{rnns}{RNNs}{Recurrent Neural Networks}
\newacronym{rnn}{RNN}{Recurrent Neural Network}
\newacronym{rsc}{RSC}{recursive systematic convolutional}
\newglossaryentry{ram}{name=RAM,description=random access memory,first=random access memory (RAM),firstplural=random access memories (RAMs)}
\newglossaryentry{rom}{name=ROM,description=read only memory,first=read only memory (ROM),firstplural=read only memories (ROMs)}
\newacronym{rms}{RMS}{recognition, mining, and synthesis}
\newacronym{rs}{RS}{reed-solomon code}
\newacronym{rtl}{RTL}{register transfer level}
\newacronym{rts}{RTS}{repeated tree search}
\newacronym{rap}{RAP}{Resilience Articulation Point}

%================= S ==================
\newacronym{snr}{SNR}{Signal-to-Noise Ratio}
\newglossaryentry{sram}{name=SRAM,description=static random access memory,first=static random access memory (SRAM),firstplural=static random access memories (SRAMs)}
\newacronym{smap}{SMAP}{serial MAP}
\newacronym{sdr}{SDR}{software-defined radio}
\newacronymf{sdram}{SDRAM}{synchronous dynamic random access memory}
\newacronym{set}{SET}{single-event transient}
\newacronym{seu}{SEU}{single-event upset}
\newacronym{siso}{SISO}{soft-input soft-output}
\newacronym{siso2}{SISO}{single-input single-output}
\newacronym{soc}{SoC}{Systems-on-Chip}
\newacronym{sova}{SOVA}{soft-output Viterbi algorithm}
%\newacronym{sram}{SRAM}{static random access memory}
\newacronym{sw}{SW}{software}
\newacronym{sm}{SM}{spatial multiplexing}
\newacronym{stbc}{STBC}{space-time block code}
\newacronym{sttc}{STTC}{space-time trellis code}
\newacronym{sic}{SIC}{successive interference cancellation}
\newacronym{sts}{STS}{single tree search}
\newacronymf{spp}{SPP}{Schwerpunktprogramm}
\newacronym{se}{SE}{Schnorr-Euchner}
\newacronym{synb}{SYN}{Syndrome-Based}

%================= T ==================
\newacronym{tmr}{TMR}{triple modular redundancy}
\newacronym{tc}{TC}{turbo code}
\newacronym{ts}{TS}{Tuple search}

%================= U ==================
\newacronymf{umts}{UMTS}{universal mobile telecommunications system}
\newacronymf{uwb}{UWB}{ultra-wide band}
\newacronym{usb}{USB}{universal serial bus}
\newacronym{umic}{UMIC}{ultra high speed mobile information and communication}

%================= V ==================
\newacronym{vn}{VN}{Variable Node}
\newacronym{vfu}{VFU}{variable node functional unit}
\newacronym{vhdl}{VHDL}{very high speed integrated circuits hardware description language}
\newacronym{vliw}{VLIW}{very large instruction word}
\newacronym{vnb}{VNB}{variable node block (in \gls{LDPC} decoder architectures)}
\newacronym{vnp}{VNP}{variable node processors (in \gls{LDPC} decoder architectures)}
\newacronym{vblast}{V-BLAST}{Vertical Bell Labs layered space time}

%================= W ==================
\newacronymf{wimax}{WiMAX}{worldwide interoperability for microwave access \cite{ieworldwide}}
\newacronymf{wimedia}{WiMedia}{name of an alliance for standardization}
\newacronymf{wifi}{WiFi}{trademark of Wi-Fi alliance}
\newacronymf{wc}{WC}{worst case, 1.1V, 125$^\circ$C}

%================= X ==================
\newglossaryentry{xmap}{name=XMAP,description={name of a pipelined MAP decoder},first=XMAP,firstplural=XMAPs}
\newacronym{xor}{XOR}{exclusive OR function}

%================= Y ==================

%================= Z ==================
\newacronym{zf}{ZF}{zero-forcing}

%% file: tex/copyrightnotice.tex
\onecolumn

{\Huge IEEE Copyright Notice}
\\
\\
{\Large \textcopyright 2018 IEEE. Personal use of this material is permitted. Permission from IEEE must be obtained for all other uses, in any current or future media, including reprinting/republishing this material for advertising or promotional purposes, creating new collective works, for resale or redistribution to servers or lists, or reuse of any copyrighted component of this work in other works.}
\\
\\
\textit{Accepted for publication, 28th International Conference on Field Programmable Logic and Applications (FPL), August, 2018, Dublin, Ireland.}

\twocolumn

%DOI: \href{<https://www.ieee.org/>}{<DOI No.>}

%% file: tex/abstract.tex
\begin{abstract}
It is well known that many types of artificial neural networks, including recurrent networks, can achieve a high classification accuracy even with low-precision weights and activations. The reduction in precision generally yields much more efficient hardware implementations in regards to hardware cost, memory requirements, energy, and achievable throughput.
In this paper, we present the first systematic exploration of this design space as a function of precision for Bidirectional Long Short-Term Memory (BiLSTM) neural network. Specifically, we include an in-depth investigation of precision vs. accuracy using a
fully hardware-aware training flow, where during training quantization of all aspects of the network including weights,
input, output and in-memory cell activations are taken into consideration.
In addition, hardware resource cost, power consumption and throughput scalability are explored as a function of precision for FPGA-based implementations of BiLSTM, and multiple approaches of parallelizing the hardware.
We provide the first open source HLS library extension of FINN \cite{umuroglu2017finn} for parameterizable hardware architectures of LSTM layers on FPGAs which offers full precision flexibility and allows for parameterizable performance scaling offering different levels of parallelism within the architecture.
Based on this library, we present an FPGA-based accelerator for BiLSTM neural network designed for optical character recognition, along with numerous other experimental proof points for a Zynq UltraScale+ XCZU7EV MPSoC within the given design space.

\end{abstract}

%% file: tex/introduction.tex
\section{Introduction}
\label{sec:intro}

\glsunset{nn}
\glsunset{rnn}
\glsunset{cnn}

\gls{rnns} and in particular \gls{lstm} have achieved state-of-the-art classification accuracy in many applications such as language modeling \cite{mikolov2010recurrent}, machine translation %\cite{sutskever2014sequence}, \cite{wu2016google},
\cite{chung2016character}, speech recognition %\cite{graves2013speech}, \cite{amodei2016deep},
\cite{chan2016listen}, and image caption generation \cite{vinyals2015show}. However, high classification accuracy comes at high compute, storage, and memory bandwidth requirements, which makes their deployment particularly challenging, especially for energy-constrained platforms such as for portable devices. 
Furthermore, many applications have hard real-time constraints such as mobile robots, hearing aids and autonomous vehicles.
%real-time applications, e.g., on-line machine translation and speech recognition running even on powerful servers. Furthermore, sequential processing becomes a problem for other latency-critical applications like autonomous vehicles and mobile robots that operate under hard real-time constrains.

Compared to feed-forward \gls{nns}, \gls{lstm} networks are especially challenging as they require state keeping in between processing steps. This has various adversary effects. First of all, extra processing is required of the recurrent connections along with the ``feed-forward'' input activations. Additionally, even though the required state memory is not particularly large, the state keeping creates data dependencies to previous steps, which forces sequentialization of the processing in parts and limits the degrees of parallelism that can be leveraged in customizable architectures and that makes multi-core general-purpose computing platforms inefficient. 
%There is a lot of ongoing research on developing networks with lower computation cost and storage consumption without impairing classification accuracy. 
Many techniques have been proposed to alleviate the compute and storage challenges described above. Among the most common ones are pruning \cite{liu2015sparse, han2015learning, wen2016learning}, and quantization (see \cref{sec:related_work} for details) or a combination thereof \cite{han2015deep}. All of them are based on the typically inherent redundancy contained within the \gls{nns}, meaning that the number of parameters and precision of operations can be significantly reduced without affecting accuracy. Within this paper, we focus on the latter method, namely reducing compute cost and storage requirements through reduction of precision in the leveraged data types, whereby we leverage Field-Programmable Gate Arrays (FPGAs) as they are they only computing platform that allows for customization of pipelined compute data paths and memory subsystems at the level of data type precision, in a programmable manner. As such they can take maximum advantage of the proposed optimization techniques.

%Basically, multiplication of parameters (weights) and activations (inputs) are the main operations of \gls{nns}. Pruning results in less multiplications and reduced storage consumption because of fewer parameters, however at a cost of irregular parallelism, more sophisticated hardware and more complicated sparse model storage representation. Quantization saves memory, and at the same time reduces computation cost because of low-precision multiplication without the drawbacks of the aforementioned approach. Additionally, the aforementioned methods reduce memory bandwidth and increase power efficiency mostly because memory accesses typically consume more energy compare to arithmetic operations and memory access costs increases with memory size. In the following, we focus on quantization-based methods. 
%As low power and low latency become essential, power hungry general-purpose computing platforms fail to meet these requirements. An efficient alternative are Field-Programmable Gate Arrays (FPGAs). FPGAs are very flexible reconfigurable computing platforms that can be optimized for a dedicated application. The main advantages compared to general-purpose computing platforms are 1) customized memory hierarchy 2) custom precision data types 3) flexible pipelined datapath and, 4) infinite bit-level parallelism. Using these advantages, FPGAs can be optimally tuned to provide ad hoc solutions to facilitate computationally intensive, time critical tasks at low power consumption.
Within this paper, we extend a vast body of existing research on implementation of quantized \gls{nn} accelerators for standard CNNs \cite{sze2017efficient} to recurrent models. 
The novel contributions are:
\begin{itemize}	
\item We conduct the first systematic exploration of the design space comprised of hardware computation cost, storage cost, power and throughput scalability as a function of precision for \gls{lstm} and \gls{bilstm} in particular.
%\item XXX We explore different choices for parallelism within the hardware for scaling performance.
\item We investigate the effects of quantization of weights, input, output, recurrent, and in-memory cell activations during training. 

\item We cross-correlate achievable performance with accuracy for a range of precisions to identify optimal performance-accuracy trade-offs with the design space.
    %\item In addition, a throughput, hardware resource cost and power consumption function for the FPGA-based implementation is derived as a function of precision. %Unlike other publications that mostly give qualitative and analytical estimations of advantages of using quantization or quantitative with respect to general computing platforms only, we quantify the savings in resource utilization, gain in throughput and power savings using actual FPGA-based hardware implementations.
    
\item To the best of our knowledge, we present the first hardware implementation of binarized (weights and activations constrained to 1 and 0) \gls{lstm} and \gls{bilstm} in particular.

\item Last but not least, we provide the first open source HLS library extension of FINN \cite{umuroglu2017finn} for parameterizable hardware architectures of \gls{lstm} layers on FPGAs, which provides full precision flexibility, allows for parameterizable performance scaling and supports a full training flow that allows to train the network on new datasets.
\end{itemize}

The paper is structured as follows: in \cref{sec:related_work} we review existing research on quantized training and FPGA-based implementations of \gls{rnns}.
In \cref{sec:theory_lstm} we review the \gls{lstm} algorithm.  We describe the training setup and procedure in Torch and PyTorch in \cref{sec:training}. \cref{sec:inference} provides the details of the hardware design. 
Finally, \cref{sec:results} presents the results and \Cref{sec:conclusion} concludes the paper.

% It is well known that many types of artificial neural networks, including recurrent networks, can achieve a high classification accuracy even with low-precision weights and activations. The reduction in precision generally yields much more efficient hardware implementations in regards to hardware cost, energy, memory requirements and achievable throughput.
% In this paper, we present a first systematic exploration of this design space as a function of precision for Bidirectional Long Short-Term Memory (BLSTM) neural network. Specifically, we include an in-depth investigation of precision vs. accuracy using a
% fully hardware-aware training flow, where during training quantization of all aspects of the network including weights,
% input, output and in-memory cell activations are taken into consideration.
% In addition, a throughput, hardware resource cost and power consumption function for the FPGA-based implementation is derived as a function of precision for fixed scaling factor and target frequency. We provide the first open source HLS library extension of FINN for parameterizable hardware architectures of LSTM layers on FPGAs which provides full precision flexibility and allows for parameterizable performance scaling.
% Based on this library, we present an FPGA-based accelerator for BLSTM neural network designed for optical character recognition that achieves the highest throughput of XXX GOPS with respect to state-of-the-art, along with numerous other experimental proof points for a Zynq UltraScale+ XCZU7EV MPSoC within the give design space.

%% file: tex/related_works.tex
\section{Related work}
\label{sec:related_work}

\subsection{Quantized Neural Networks}
% \textcolor{red}{We need to list publications that justify research with respect to binarized and multi-bit quantized neural networks. With quantized weights and activations. Probably, we need to list only the latest works.} 

The most extreme quantization approach is binarization that results in Binarized Neural Networks (BNNs) with binarized weights only \cite{courbariaux2015binaryconnect} or with both weights and activations quantized to 1-bit (usually \{-1,1\} during training and \{0,1\} at inference) that was firstly proposed in \cite{hubara2016binarized}, \cite{courbariaux2016binarized}. Compared with the 32-bit full-precision counterpart, binarized weights drastically reduce memory size and accesses. At the same time, BNNs significantly reduce the complexity of hardware by replacing costly arithmetic operations between real-value weights and activations with cheap bit-wise XNOR and bitcount operations, which altogether leads to much acceleration and great increase in power efficiency.
It has been shown that even 1-bit binarization can achieve reasonably good performance in some applications. In \cite{rastegari2016xnor} they proposed to use a real-value scaling factor to compensate classification accuracy reduction due to binarization and achieved better performance than pure binarization in \cite{hubara2016binarized}. However, binarization of both weights and activations can lead to undesirable and non-acceptable accuracy reduction in some applications compared with the full-precision networks. To bridge this gap, recent works employ quantization with more bits like ternary \{-1,0,1\} \cite{li2016ternary, zhu2016trained} and low bit-width \cite{hubara2016quantized, zhou2016dorefa, zhou2017balanced} networks that achieve better performance bringing a useful trade-off between implementation cost and accuracy. %In \cite{hubara2016quantized}, \cite{zhou2016dorefa} they have shown that using multi-bit activations, while keeping 1-bit weights, improves accuracy approaching the accuracy level of full single-precision floating point models.  
%In this paper, we continue the research within this design space with a focus on LSTM NNs. 

Among all existing works on quantization and compression, most of them focus on \gls{cnns} while less attention has been paid to \gls{rnns}. The quantization of the latter is even more challenging. \cite{hubara2016quantized}, \cite{zhou2017balanced}, \cite{he2016effective} have shown high potential for networks with quantized multi-bit weights and activations. %showed that quantization of \gls{lstm} with less than 4-bits weights and activations, results in a noticeable gap compared to those with full-precision.
Recently, \cite{hou2016loss} showed that \gls{lstm} with binarized weights only can even surpass full-precision counterparts. \cite{xu2018alternating} has addressed the problem by quantizing both weights and activations. They formulate the quantization as an optimization problem and solve it by the binary search tree. Using language models and image classification task, they have demonstrated that with 2-bit quantization of both weights and activations, the model suffers only a modest loss in the accuracy, while 3-bit quantization was on par with the original 32-bit floating point model.

%\cite{xu2018alternating} addresses the problem by quantizing the network, both weights and activations. They formulate the quantization as an optimization problem and solve it by the binary search tree.  The quantization was tested for two well-known RNNs, i.e., LSTM and GRU, on the language models and image classification task and demonstrated that with 2-bit quantization of both weights and activations, the model suffers only a modest loss in the accuracy, while 3-bit quantization was on par with the original 32-bit floating point model.

Unlike other publications that mostly give qualitative and analytical estimations of advantages of using quantization with respect to general-purpose computing platforms only, we quantify the savings in resource utilization, gain in throughput and power savings using actual FPGA-based hardware implementations. %We explore classification accuracy as a function of precision in more details than in other research by considering quantization of all aspects of the network including weights, input, output and in-memory cell activations are taken into consideration.
%We also propose a new approach of quantizing activations of \gls{rnns}. 
Furthermore, we propose a new approach when the recurrent and output activations are quantized differently that keeps resource utilization of a hidden layer, while providing higher accuracy with respect to uniform approach. %with lower bit-width.
To the best of our knowledge, we are the first investigating influence of precision on classification accuracy on \gls{ocr} using fully hardware-aware training flow, where during training quantization of all aspects of the \gls{lstm} network including weights, input, output, recurrent, and in-memory cell activations are taken into consideration.

%Summarizing, 3-bit quantization of weights and activations should result in a model that has the same accuracy or even surpasses a full-precision model. Quantization with binarized weights and multi-bit activations can result in insignificant loss in classification accuracy. 

%\cite{hubara2016quantized}, \cite{zhou2017balanced} showed promising results of multi-bit quantized \gls{rnns} for language modeling. Although using less than 4-bits weights and activations, the results with quantization still had noticeable gap with those with full-precision.

\subsection{Recurrent Neural Networks on FPGA}

\subsubsection{With Quantization}~\\

In \cite{chang2015recurrent}, authors presented an FPGA-based hardware implementation of pre-trained \gls{lstm} for character level language modeling. They used 16-bit quantization for weights and input data, both of which were stored in off-chip memory. Intermediate results were also sent back to off-chip memory in order to be read for computing next layer or next timestep output that has been identified as a performance bottleneck. %Implemented on Xilinx Zynq XC7Z020, the design could reach a throughput of 0.284 GOPS.
The following works tried to de-intensify off-chip memory communication by storing a model and intermediate results on chip. An \gls{lstm} network for a learning problem of adding two 8-bit numbers has been implemented in \cite{ferreira2016fpga}. The design could achieve higher throughput than \cite{chang2015recurrent}, while using 18-bit precision for activations and weights but stored in on-chip memory. %The design could achieve 4.53 GOPS using 18-bit precision for activations and weights stored on-chip.
\cite{lee2016fpga} proposed a real-time, low-power FPGA-based speech-recognition system with higher energy efficiency than GPU. The FPGA-based implementation of \gls{lstm} used 8-bit activations and 6-bit weights stored in on-chip memory. They also experimented with 4-bit weights, however with significant accuracy degradation.
\cite{guan2017fpga} implemented an \gls{lstm} network for the speech recognition application. Although using higher precision 32-bit floating-point weights stored in off-chip memory, they could achieve higher throughput than previous works, because of smart memory organization with on-chip ping-pong buffers to overlap computations with data transfers. %They used on-chip ping-pong buffers to overlap computations with data transfers. The design could achieve a peak performance of 7.26 GFLOPS. 
%A work presented in \cite{hao2017implementation} achieved a throughput of 20 GOPS implementing a fixed-point \gls{lstm} network on Xilinx PYNQ XC7Z020 FPGA for language modeling. 
FP-DNN - an end-to-end framework proposed in \cite{guan2017fp} that takes TensorFlow-described network models as input, and automatically generates hardware implementations of various \gls{nns} including \gls{lstm} on FPGA using RTL-HLS hybrid templates. They used RTL for implementing matrix multiplication parts and to express fine-grained optimizations, while HLS to implement control logic. The FPGA implementation made use of off-chip memory for storing 16-bit fixed-point format weights and intermediate results, and ping-pong double buffers to overlap communication with computation. %Implemented on Catapult system with Altera Stratix-V GSMD5 FPGAs, the design could achieve 315.85 GOPS for \gls{lstm} in language modeling task.

The current work is based on existing architecture proposed in \cite{rybalkin2017hardware}, where the authors presented the first hardware architecture designed for \gls{bilstm} for \gls{ocr}. The architecture was implemented with 5-bit fixed-point numbers for weights and activations that allowed to store all intermediate results and parameters in on-chip memory. %Implemented on Xilinx Zynq XC7Z045, the design could provide 693.12 GOPS. %They could show acceleration over Intel Xeon E5 processor, at the same time being more energy efficient than embedded processors.
The current research, differs from the previous works by combining both state-of-the-art quantization techniques in training and advanced hardware architecture. Non of the mentioned works targeting hardware implementation was addressing efficient hardware-aware training. The architecture is on-chip memory centric that avoids using off-chip memory for storing weights like in  \cite{chang2015recurrent}, \cite{guan2017fpga}. Efficient quantization at training allows for efficient on-chip implementation even larger networks because of reduced precision without effecting accuracy rather than using off-chip bandwidth and higher precision like in \cite{guan2017fp}. Non of the previous works have demonstrated efficient support for a wide range of precisions. %We provide the first open source HLS library extension of FINN \cite{umuroglu2017finn} for parameterizable hardware architectures of \gls{lstm} layers on FPGAs which offers full precision flexibility and allows for parameterizable performance scaling offering different levels of parallelism within the architecture.
We provide an open source HLS library that supports a wide scope of precisions of weights and activations ranging from multi-bit to fully binarized version. To the best of our knowledge, we preset the first fully binarized implementation of \gls{lstm} \gls{nn} in hardware. Finally, the library is modular and optimized, providing customized functions to the various layer types rather than only matrix multiplication functions.      
In contrast to \cite{guan2017fp}, FINN hardware library is purely implemented using HLS rather than RTL-HLS hybrid modules. 
\\

% In \cite{rybalkin2017hardware}, the authors presented the first hardware architecture designed for \gls{bilstm} for \gls{ocr}. The architecture was implemented with 5-bit fixed-point numbers for weights and activations that allowed to store all intermediate results and parameters in on-chip memory. Implemented on Xilinx Zynq XC7Z045, the design could provide 693.12 GOPS. They could show acceleration over Intel Xeon E5 processor, at the same time being more energy efficient than embedded processors.

\subsubsection{With Pruning and Quantization}~\\

%\cite{rizakis2018approximate} addressed implementation of \gls{lstm} \gls{nn} at a constrained time budget by introducing an end-to-end framework supporting an iterative approximation method that combines low-rank compression and pruning, while using single-precision floating-point format for computations. Their method does not require retraining phase. Proposed methods required up to 6.5 time less time to achieve the same application-level accuracy compared to a baseline without any compression.

For the sake of completeness, we also mention architectures targeting sparse models unlike our design that supports dense models. A design framework for a hybrid \gls{nn} (\gls{cnn} + \gls{lstm}) that uses configurable IPs together with a design space exploration engine was presented in \cite{zhang2017high}. The proposed HLS IPs implement MAC operations. They used pruning and 12-bit weights and 16-bit fixed-point activations. ESE \cite{han2017ese} presented an efficient speech recognition engine implementing \gls{lstm} on FPGA that combined both pruning and 12-bit compact data types. The design can operate on both dense and sparse models. 
\cite{wang2018c} proposed a compression technique for \gls{lstm} that reduces not only the model size, but also eliminates irregularities of compression and memory accesses. The datapath and activations were quantized to 16 bits. They also presented a framework called C-LSTM to automatically map \gls{lstm} onto FPGAs.

%Unlike \cite{rizakis2018approximate}, 
Our approach is based on retrained optimization based on quantization that avoids a use of redundant high precision calculations. In contrast to aforementioned works our architecture is on-chip memory oriented and designed to operate on dense models. Architecture presented  in \cite{wang2018c} is designed to operate on compressed weight matrices with block-circulant matrix based structured compression technique and cannot process not compressed models. Non of those works explored very low bit-width weights and activations.

%% file: tex/theory.tex
\section{\gls{lstm} Theory}
\label{sec:theory_lstm}

For the sake of clarity, we review the basic \gls{lstm} algorithm, earlier presented in \cite{hochreiter1997long}, \cite{graves2012supervised}.

\begin{equation}
	\label{eq:lstm}
    \begin{split}
    	&\boldsymbol{I}^t =  l(
        					  \boldsymbol{W}_I \boldsymbol{x}^t + 
                              \boldsymbol{R}_I \boldsymbol{y}^{t-1} + 
                              \boldsymbol{b}_I
			        		 )
		\\
    	&\boldsymbol{i}^t =  \sigma(
                                   \boldsymbol{W}_i \boldsymbol{x}^t + 
                                   \boldsymbol{R}_i \boldsymbol{y}^{t-1} + 
                                   \boldsymbol{p}_i \odot \boldsymbol{c}^{t-1} +
                                   \boldsymbol{b}_i
			        			  )
        \\
    	&\boldsymbol{f}^t =  \sigma(
                                   \boldsymbol{W}_f \boldsymbol{x}^t + 
                                   \boldsymbol{R}_f \boldsymbol{y}^{t-1} + 
                                   \boldsymbol{p}_f \odot \boldsymbol{c}^{t-1} +
                                   \boldsymbol{b}_f
				                  )
        \\
    	&\boldsymbol{c}^t =  \boldsymbol{i}^t \odot \boldsymbol{I}^t +
        					\boldsymbol{f}^t \odot \boldsymbol{c}^{t-1}
        \\
    	&\boldsymbol{o}^t =  \sigma(
                                   \boldsymbol{W}_o \boldsymbol{x}^t + 
                                   \boldsymbol{R}_o \boldsymbol{y}^{t-1} + 
                                   \boldsymbol{p}_o \odot \boldsymbol{c}^{t} +
                                   \boldsymbol{b}_o
				                  )
        \\
    	&\boldsymbol{y}^t =  \boldsymbol{o}^t \odot h(\boldsymbol{c}^t)
\end{split}
\end{equation}

The LSTM architecture is composed of a number of recurrently connected ``memory cells''.
Each memory cell is composed of three multiplicative gating connections, namely input, forget, and output gates ($\boldsymbol{i}$, $\boldsymbol{f}$, and $\boldsymbol{o}$, in Eq. \ref{eq:lstm}); the function of each gate can be interpreted as write, reset, and read operations, with respect to the cell internal state ($\boldsymbol{c}$). The gate units in a memory cell facilitate the preservation and access of the cell internal state over long periods of time. The peephole connections ($\boldsymbol{p}$) are supposed to inform the gates of the cell about its internal state. However, they can be optional and according to some research even redundant \cite{breuel2015benchmarking}. There are recurrent connections from the cell output ($\boldsymbol{y}$) to the cell input ($\boldsymbol{I}$) and the three gates. Eq. \ref{eq:lstm} summarizes formulas for \gls{lstm} network forward pass. Rectangular input weight matrices are shown by $\boldsymbol{W}$ , and square weight matrices by $\boldsymbol{R}$. $\boldsymbol{x}$ is the input vector, $\boldsymbol{b}$ refers to bias vectors, and t denotes the time (and so t-1 refers to the previous timestep). Activation functions are point-wise non-linear functions, that is \textit{logistic sigmoid} ($\frac{1}{1+e^{-x}}$) for the gates ($\sigma$) and \textit{hyperbolic tangent} for input to and output from the node ($l$ and $h$). Point-wise vector multiplication is shown by $\odot$ (equations adapted from \cite{greff2017lstm}).

Bidirectional \gls{rnn}s were proposed to take into account impact from both the past and the future status of the input signal by presenting the input signal forward and backward to two separate hidden layers both of which are connected to a common output layer that improves overall accuracy \cite{graves2005framewise}.

% \gls{ctc} is the output layer used with \gls{lstm} networks designed for sequence labelling tasks. Using the \gls{ctc} layer, \gls{lstm} networks can perform the transcription task without requiring pre-segmented input, as it is trained to predict the conditional probabilities of the possible output labellings, given input sequences. \gls{ctc} layer has $K$ units, $K-1$ being the number of elements in the alphabet of labels. The outputs are normalized using softmax activation function at each timestep. The softmax activation function ensures that network outputs are all in the range between zero and one, and they sum to one at each timestep. This means that the outputs can be interpreted as the probabilities of the characters at a given timestep (column in our case). The first $K-1$ outputs provide predictions over the probabilities of their corresponding labels for the input timestep, and the remaining one unit estimates the probability of a blank or no label. Summation over the probabilities that correspond to a particular output label yields its estimated probability (see \cite{graves2006connectionist} for more details and related
% equations).

\gls{ctc} is the output layer used with \gls{lstm} networks designed for sequence labelling tasks. Using the \gls{ctc} layer, \gls{lstm} networks can perform the transcription task without requiring pre-segmented input (see \cite{graves2006connectionist} for more details and related equations).

%% file: tex/training.tex
\section{Training}
\label{sec:training}

\subsection{Topology}

For our experiments, we implemented a quantized version of \gls{bilstm} \gls{nn} on Torch7 \cite{torch7} and PyTorch \cite{pytorch} frameworks. For Torch, our Quantized \gls{bilstm} (QBiLSTM) is based on \gls{rnn} library \cite{leonard2015rnn} that provides full-precision implementation of various \gls{rnn} models. For PyTorch, our quantized implementation, called \textit{pytorch-quantization}, is based on official full-precision ATen \cite{aten} accelerated PyTorch BiLSTM implementation, on top of which we implemented an OCR training pipeline, called \textit{pytorch-ocr}. Both \textit{pytorch-quantization} \cite{pytorch-quantization} and \textit{pytorch-ocr} \cite{pytorch-ocr} are open-source. 

\begin{table}[H]
	\centering
	\caption{Training. General information.}
	\label{info}
	\begin{tabular}{ll}
		\toprule
		Number of images in training set, $\boldsymbol{T}_{train}$           	& 400 \\
        Number of images in test set, $\boldsymbol{T}_{test}$           		& 100 \\
        Number of pixels per column (Input layer), $\boldsymbol{I}$ 		    & 32  \\
		Max number of columns per image, $\boldsymbol{C}$             			& 732 \\		
		Number of \gls{lstm} memory cells in a hidden layer, $\boldsymbol{H}$	& 128 \\		
		Number of units in the output layer, $\boldsymbol{K}$ 					& 82  \\
        Minibatch size, $\boldsymbol{B}$ 										& 32  \\
        Maximum number of epochs, $\boldsymbol{E}$ 								& 4000  \\
        Learning rate, $\boldsymbol{L}$ 										& 1e-4  \\
		\bottomrule
	\end{tabular}
\end{table}

We use \gls{ocr} plain-text dataset of $\boldsymbol{T}_{train}$ images for training and $\boldsymbol{T}_{test}$ images for test. Each sample is a gray-scale image of a text-line with a fixed height of $\boldsymbol{I}$ pixels and width up to $\boldsymbol{C}$ pixels. In \gls{lstm} terminology, height and width corresponds, respectively, to the input size of the \gls{lstm} cell and to the length of the input sequence.

\begin{figure}[]
	\centering
	\includegraphics[width=0.9\columnwidth]{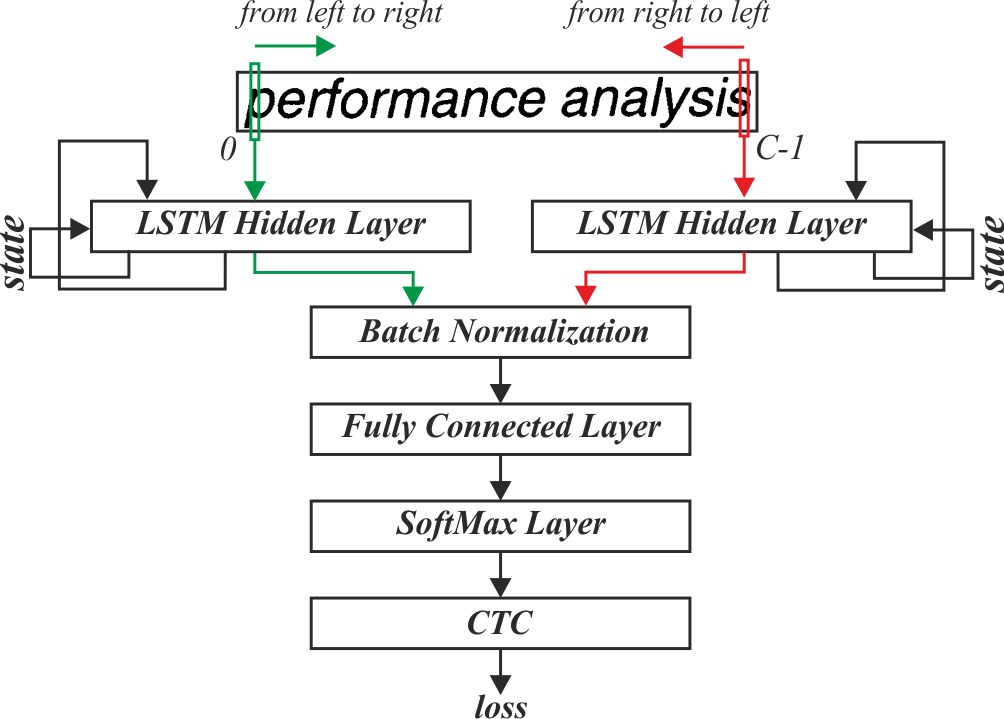}
	\caption{Network topology used during training. Only forward path is shown.}
 	\label{fig:training_topology}
 \end{figure}

In all experiments, we adopt the following topology:
\begin{itemize}
\item An input layer that feeds images column by column along the width dimension $\boldsymbol{C}$.
\item A single \gls{bilstm} layer, composed of two sub-layers, each comprising of a distinct set of $\boldsymbol{H}$ \gls{lstm} cells without peepholes. The two sub-layers go through the input image, respectively, from left-to-right and from right-to-left, each generating an output sub-sequence of length $\boldsymbol{C}$ and feature size $\boldsymbol{H}$. At each timestep, each gate of each memory cell operates on $1 + \boldsymbol{I} +\boldsymbol{H}$ values, comprising of a single bias value, $\boldsymbol{I}$ pixels and the $\boldsymbol{H}$ output values computed from the previous column in the sequence, according to Eq. \ref{eq:lstm}. The \gls{bilstm} layer's output sequence is composed of the right-to-left output sub-sequence taken in reverse order and concatenated with the left-to-right output sub-sequence along the feature dimension.
\item A batch normalization step with minibatch of size $\boldsymbol{B}$ on the output sequence of the \gls{bilstm} layer, to speed up the training.
\item A fully connected layer (with dropout 0.2 in Torch) mapping each column of the normalized output sequence to a vector of $\boldsymbol{K}$ features, one for each of the symbols in the dataset's alphabet including a blank. 
\item A softmax layer, to interpret the output of the fully connected layer as a vector of probabilities over the alphabet.
\item A \gls{ctc} layer as loss function, using a GPU-accelerated implementation from Baidu Research \cite{baidu_research}.
\end{itemize}

All training experiments were running on GPUs. After running a series of experiments, we take the best-effort validation accuracy over the multiple experiments and over the training frameworks (Torch and PyTorch) to evaluate the efficiency of the configurations. The classification accuracy is based on Character Error Rate (CER) computed as the Levenshtein distance \cite{wagner1974string} between a decoded sequence and a ground truth. In \cref{sec:results}, we list accuracy of models trained at maximum over $\boldsymbol{E}$ epochs using a SGD learning rule with learning rate $\boldsymbol{L}$.

% In all experiments, we used the same topology: 1) $\boldsymbol{N}$ LSTM memory cells in hidden layer, 2) $\boldsymbol{K}$ neurons in output fully connected layer (with dropout 0.2 in Torch) that receive concatenated output from the BiLSTM layer, and 3) a batch normalization layer. As a loss function we used a \gls{ctc} implementation from Baidu Research \cite{baidu_research}. All training experiments were running on GPUs. We used a best-effort test accuracy on our dataset to evaluate the efficiency of the configurations. In \cref{sec:results}, we list accuracy of models trained at maximum over  $\boldsymbol{E}$ epochs using a SGD learning rule with learning rate $\boldsymbol{L}$. 

%We use Batch Normalization with a minibatch of size 100 to speed up the training. We  report  the  test  error  rate  associated with the best validation error rate after 1000 epochs. The results are reported in Table 1

% In XNOR-Net, they have shown that the optimal estimation of a binary weight filter can be simply achieved by taking the sign of weight values. The optimal scaling factor is the average of absolute weight values.

%in DoReFa-Net: wb=Ew*sign(w). Here Ew calculates the mean of absolute values of full precision weights w as layer-wise scaling factors.
%\boldsymbol{s_q} =clip(\dfrac{round(\boldsymbol{x} \cdot 2^{\boldsymbol{f}})}{2^{\boldsymbol{f}}},min,max), \text{where}

\subsection{Quantization}

%Our training approach is based on the one presented in \cite{courbariaux2016binarized}: during training, both forward and backward propagations operate on quantized weights and activations, while gradients are computed at full-precision and applied to a shadow full-precision version of the weights.

Our quantization approaches are based on those presented in \cite{rastegari2016xnor}, \cite{hubara2016quantized}, \cite{zhou2016dorefa}. For multi-bit weights and activations, we adopt Eq. \ref{eq:quantization} as a quantization function, where $\boldsymbol{x}$ is a full-precision activation or weight, $\boldsymbol{x_q}$ is the quantized version, $\boldsymbol{k}$ is a chosen bit-width, and $\boldsymbol{f}$ is a number of fraction bits. For multi-bit weights, the choice of parameters is shown in Eq. \ref{eq:quantization_signed}. For multi-bit activations, we assume that they have passed through a bounded activation function before the quantization step, which ensures that the input to the quantization function is either $\boldsymbol{x} \in [0,1)$ after passing through a \textit{logistic sigmoid} function, or $\boldsymbol{x} \in [-1,1)$ after passing through a \textit{tanh} function. As for the quantization parameters, we use Eq. \ref{eq:quantization_signed} after the \textit{tanh} and Eq. \ref{eq:quantization_unsigned} after the \textit{logistic sigmoid}.

% In our approach the weights and activations are quantized deterministically. We assume that the activations before quantization step, have been passed through a bounded activation function, which ensures that $\boldsymbol{a} \in [0,1)$ after \textit{logistic sigmoid} function, and $\boldsymbol{a} \in [-1,1)$ after \textit{tanh} function. In the case of multi-bit quantization of weights and activations passed through \textit{tanh}, we used formula Eq. \ref{eq:quantization}, where $\boldsymbol{a}$ and $\boldsymbol{w}$ are real-value activations and weights respectively, $\boldsymbol{a_q}$ and $\boldsymbol{w_q}$ are quantized versions, $\boldsymbol{k}$ bit-width of quantized number, $\boldsymbol{f}=\boldsymbol{k}-1$ number of fraction bits. 

  \begin{subequations}
      \begin{equation}\label{eq:quantization}    
    \begin{split} 
    	\boldsymbol{x_q} =clip(round(\boldsymbol{x} \cdot 2^{\boldsymbol{f}}) \cdot 2^{-\boldsymbol{f}},min,max), \text{where} 
    \end{split}
    \end{equation}

    \begin{equation}\label{eq:quantization_signed}
    \begin{aligned}
    \begin{split}         
        \boldsymbol{f} = \boldsymbol{k}-1
        \\
        min=-(2^{(\boldsymbol{k}-\boldsymbol{f}-1)})=-1
        \\
        max=-min-2^{-\boldsymbol{f}}=1-2^{-\boldsymbol{f}}   
    \end{split}
    \end{aligned}    
    \end{equation}

    \begin{equation}\label{eq:quantization_unsigned}
    \begin{split}     
        \boldsymbol{f} = \boldsymbol{k}
        \\
        min=0
        \\
        max=2^{\boldsymbol{k}-\boldsymbol{f}}-2^{-\boldsymbol{f}}=1-2^{-\boldsymbol{f}}        
    \end{split}
    \end{equation} 
  \end{subequations}

In the case of binarized activations $\boldsymbol{a_b}$, we apply a $sign$ function as the quantization function \cite{rastegari2016xnor}, as shown in Eq. \ref{eq:binarization_activations}. In the case of binarized weights $\boldsymbol{w_b}$, on top of the sign function we scale the result by a constant factor, as shown in Eq. \ref{eq:binarization_weights}.

\begin{equation} 
\label{eq:binarization_activations}
\begin{split} 
 \boldsymbol{a_b}=sign(\boldsymbol{x})
 \end{split}
\end{equation}

\vspace{-1em}

  \begin{equation} 
  \label{eq:binarization_weights}
  \begin{split}
  		\boldsymbol{w_b}=sign(\boldsymbol{x}) \cdot scaling\_factor      
      \\
      	scaling\_factor=\dfrac{1}{sqrt(\boldsymbol{H} + \boldsymbol{I})}
  \end{split}
  \end{equation}

Inspired by \cite{rastegari2016xnor}, we also experimented with using the mean of the absolute values of the full-precision weights as a layer-wise scaling factor for the binarized weights. However, it didn't show any improvements in classification accuracy with respect to the constant scaling factor described earlier. 

In our experiments, we also considered different approaches for training quantized \gls{bilstm}: In Torch, quantized model are trained from scratch. In PyTorch, quantized model are trained from scratch as well, but we leave out the quantization of the internal activations to a second short ($E = 40$) retraining step, during which we also fold the parameters of the batch-norm layer into the full-precision shadow version of the fully connected layer, forcing the quantized version to re-converge. Avoiding the overhead of computing the quantized internal activations during the main training step grants some speed-up with no impact on accuracy. We run all experiments with all in-memory cell activations and weights of the fully connected output layer quantized to 8 bits, while varying quantization of weights, input, output, and recurrent activations.

%We conducted a systematic exploration of precision as a function of accuracy of BiLSTM \gls{nn} using a fully hardware-aware training flow, where during training, quantization of all aspects of the network including weights, input, output and in-memory cell activations are taken into consideration. 

%The multi-bit quantization in Torch was performed using the following formula:

% The \gls{lstm} implemented in the paper includes an input layer that receives images column by column. In the testing set of $T$ text lines, each is represented with a gray-scale image with $P$ pixels height and variable length up to $C$ columns. The \gls{blstm} is composed of a Forward Hidden Layer (FHL) that includes $n^f_i$ \gls{lstm} cells processing images from left to right and a Backward Hidden Layer (BHL) that includes $n^b_i$ \gls{lstm} cells processing images from right to left each with a distinct set of weights, where $(i \in 0 \dots N-1)$. Each \gls{lstm} gate receives $S^H$ source values that include single bias value,
% $P$ pixels and $N$ output values received recurrently from the previous time step. The output layer includes $n^o_i$ units with $(i \in 0 \dots K-1)$, where each receives a single bias value and
% $2N$ values that are concatenated outputs of the hidden layers from corresponding columns of the image. For classification purposes, we make use of \gls{ctc} layer.

%% file: tex/inference.tex
\section{Architecture for Inference}
\label{sec:inference}

In the following, we used an existing architecture proposed in \cite{rybalkin2017hardware}. The design has been modified for parameterizable precision and parallelization, and optimized for higher frequency. The hardware accelerator is implemented on Xilinx Zynq UltraScale+ MPSoC ZCU104. The software part is running on \gls{pc} and hardware part is implemented in \gls{pl}. The software part plays an auxiliary role. It reads all images required for the experiment from SD card to DRAM and iteratively configures hardware blocks for all images. As soon as the last image has been processed, the software computes a Character Error Rate (CER) of the results based on Levenshtein distance \cite{wagner1974string} with respect to a reference.

The constructed network at inference differs from the one used during training. At inference, the batch normalization layer is merged with the fully connected layer into output layer. We avoid using a softmax activation layer and replace the \gls{ctc} with a hardware implementation of the greedy decoder that maps each output vector of probabilities to a single symbol by taking a maximum over it.

\begin{figure}[H]
	\centering
	\includegraphics[width=0.5\columnwidth]{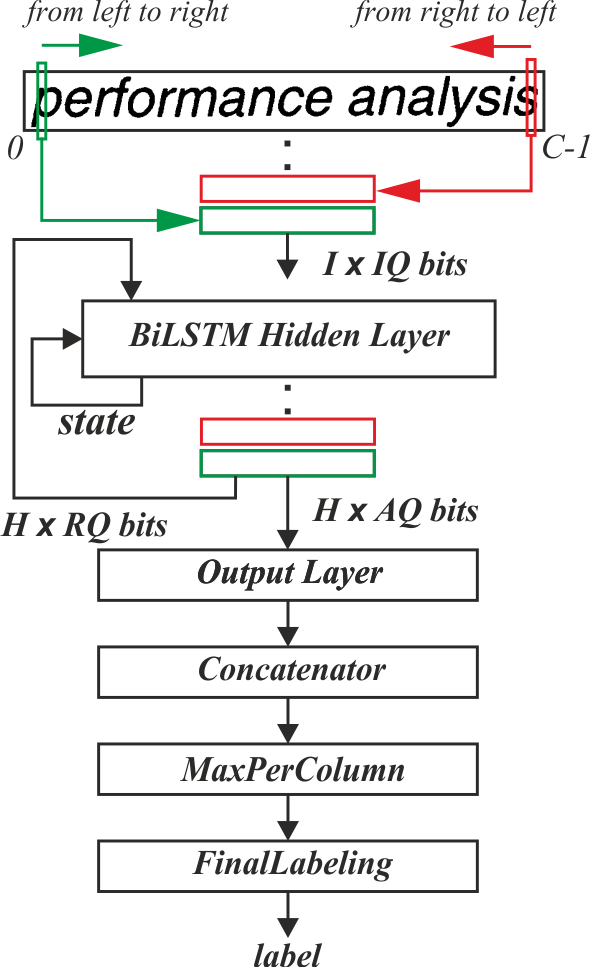}
	\caption{Network topology used during inference.}
 	\label{fig:inference_topology}
 \end{figure}

The architecture of the hidden layer processes inputs corresponding to the left-to-right and right-to-left timesteps in an interleaved fashion, which avoids a duplication of the hidden layer in the case of bidirectional network. Additionally, processing independent inputs (from different directions) in an interleaved manner allows to keep the pipeline always busy, otherwise the pipeline will be part time idle because of the recurrent dependence. 
%If the datapath processes inputs corresponding to the same direction sequentially, the pipeline will be part time idle because of the recurrent dependence.
The same approach can be applied to uni-directional \gls{rnn} that will allow for processing separate samples alike in a batch. Accordingly, the output layer is implemented to process the interleaved outputs corresponding to different directions. However, the $MaxPerColumn$ has to operate on outputs that correspond to the same timestep (column). For that the $Concatenator$ buffers the intermediate results until the first matching column that is a $C/2$ in an input sequence, and outputs the summed values that correspond to the same column but from different directions. The detailed explanation of the architecture can be found in \cite{rybalkin2017hardware}.

\subsection{Parameterizable Unrolling}
\begin{figure}[H]
	\centering
	\includegraphics[width=0.9\columnwidth]{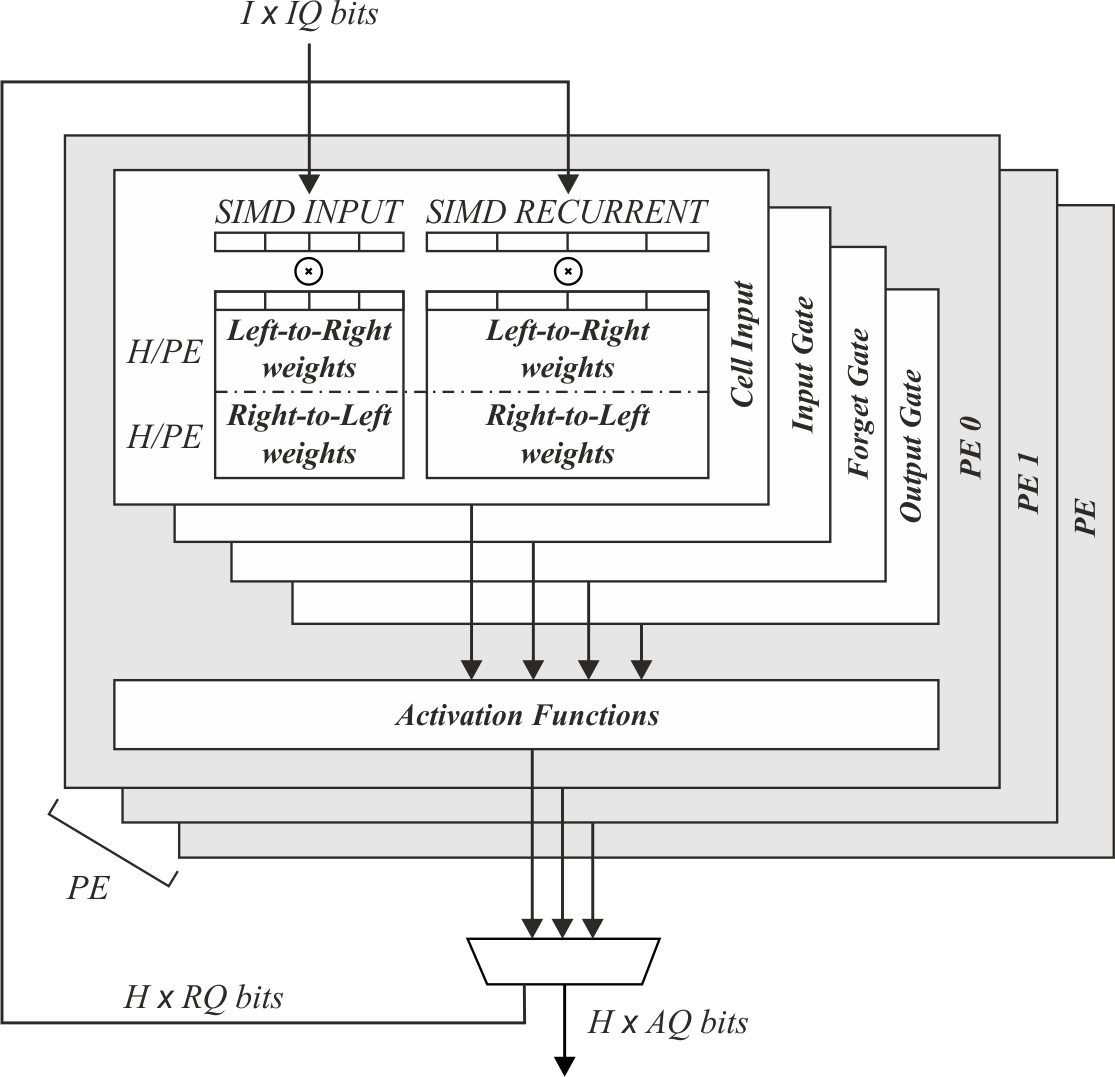}
	\caption{LSTM Cell internal structure with SIMD and PE folding}
 	\label{fig:LSTM_cell}
 \end{figure}

The parameterizable architecture allows to apply coarse-grained parallelization on a level of \gls{lstm} cells, and fine-grained parallelization on a level of dot products, as shown in Fig. \ref{fig:LSTM_cell}. The former, indicated as \texttt{PE} unrolling, allows the concurrent execution of different \gls{lstm} cells, while the latter, folds the execution of a single cell in multiple cycles. 
This flexibility allows for tailoring of parallelism according to the available resources on the target device. Increasing \texttt{PE} will increase parallelism, leading to higher hardware usage and decreased latency, while \texttt{SIMD} folding will decrease parallelism, thus scaling down hardware requirements at the expense of slower performance.
 
When \texttt{PE} unrolling is performed, the \gls{lstm} cell will be duplicated \texttt{PE} times, generating for each clock cycle \texttt{PE} output activations. When \texttt{SIMD} folding is applied, each gate will receive a portion of the input column each cycle, namely \texttt{SIMD\_INPUT} and a portion of the recurrent state \texttt{SIMD\_RECURRENT}.
Each gate will generate the output of the dot product every \texttt{F\_s} cycles, evaluated as $F\_s = \frac{I}{SIMD\_INPUT} \equiv \frac{H}{SIMD\_RECURRENT}$.

% \begin{figure*}[H]
% 	\centering
% 	\includegraphics[width=1.8\columnwidth]{./fig/Neuron_SIMD_PE.png}
% 	\caption{LSTM Cell internal structure with SIMD and PE folding}
%  	\label{fig:LSTM_cell}
%  \end{figure*}

%% file: tex/results.tex
\section{Results}
\label{sec:results}

%The following prototypes have been implemented on the Xilinx Zynq UltraScale+ MPSoC ZCU104.
The ZCU104 board features a XCZU7EV device, featuring 624 BRAMs, 96 URAMs, 1728 DSP48, 230 kLUTs and 460 kFFs. The HLS synthesis and place and route have been performed by Vivado HLS Design Suite 2017.4.

\begin{figure}[H]
	\centering
	\includegraphics[width=1.0\columnwidth]{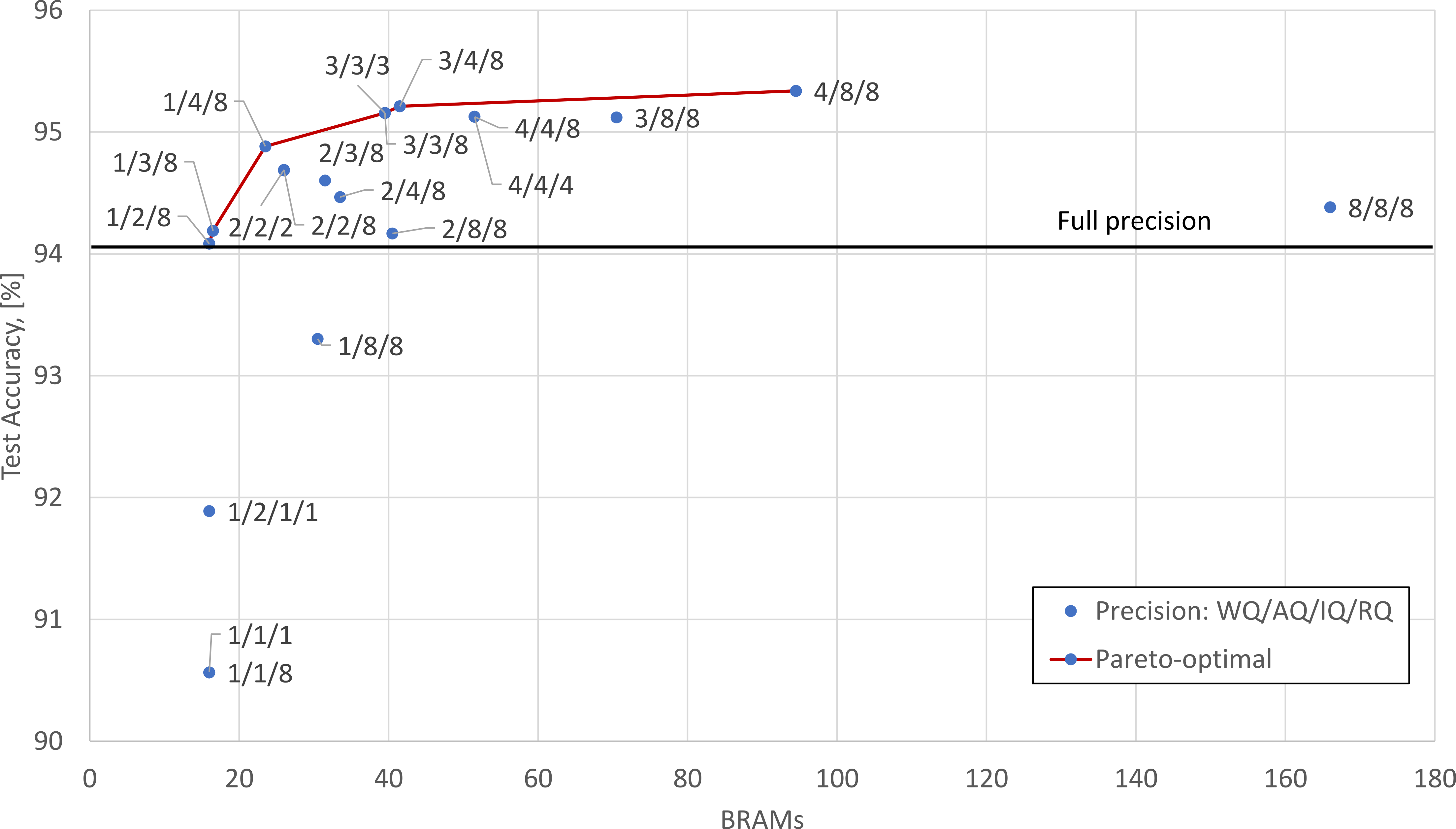}
	\caption{Pareto-frontier for Accuracy against BRAMs}
 	\label{fig:errorBRAM}
 \end{figure}

\begin{figure}[H]
	\centering
	\includegraphics[width=1.0\columnwidth]{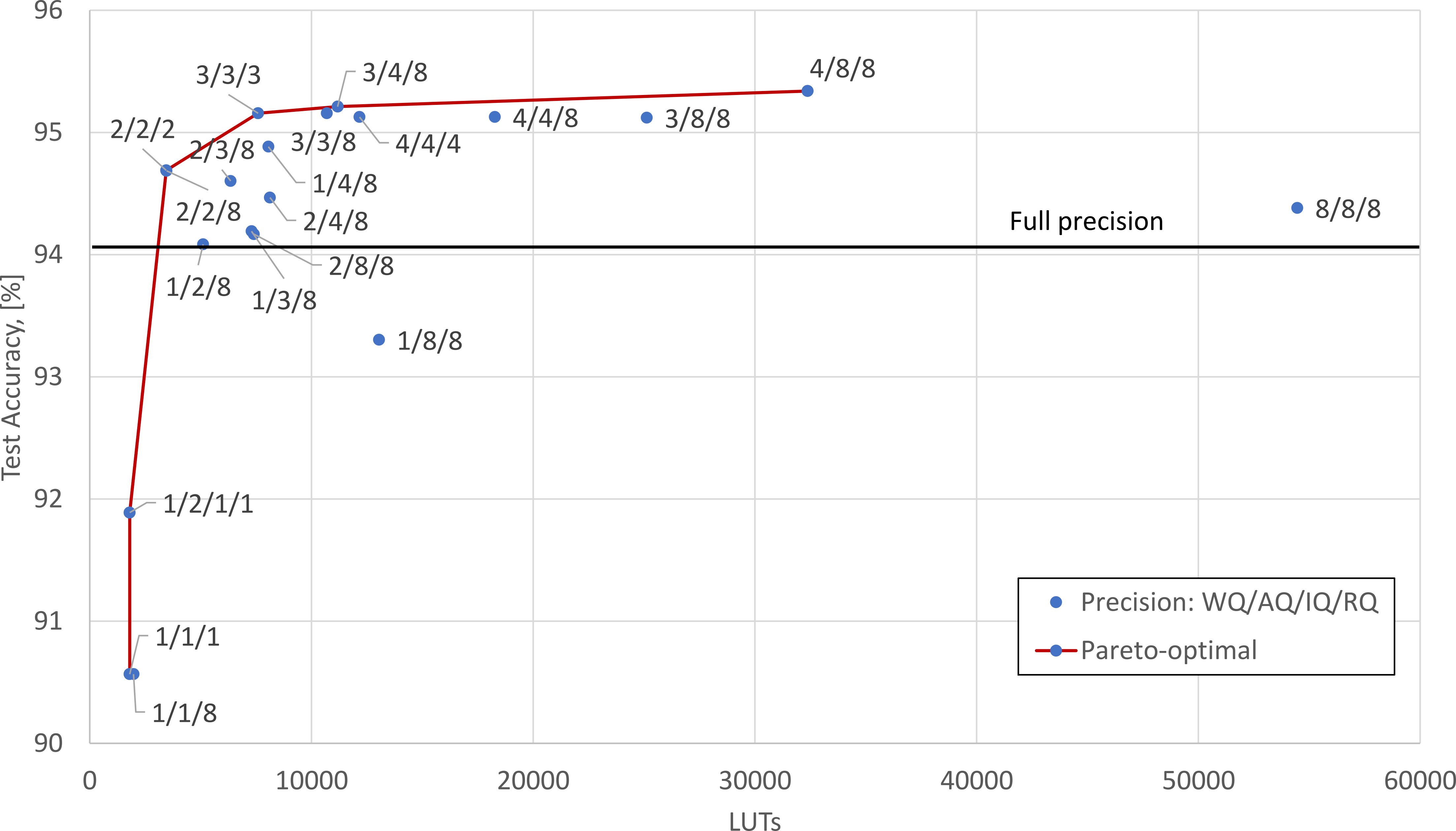}
	\caption{Pareto-frontier for Accuracy against LUTs}
 	\label{fig:errorLUT}
 \end{figure}
 
\begin{figure}[H]
	\centering
	\includegraphics[width=1.0\columnwidth]{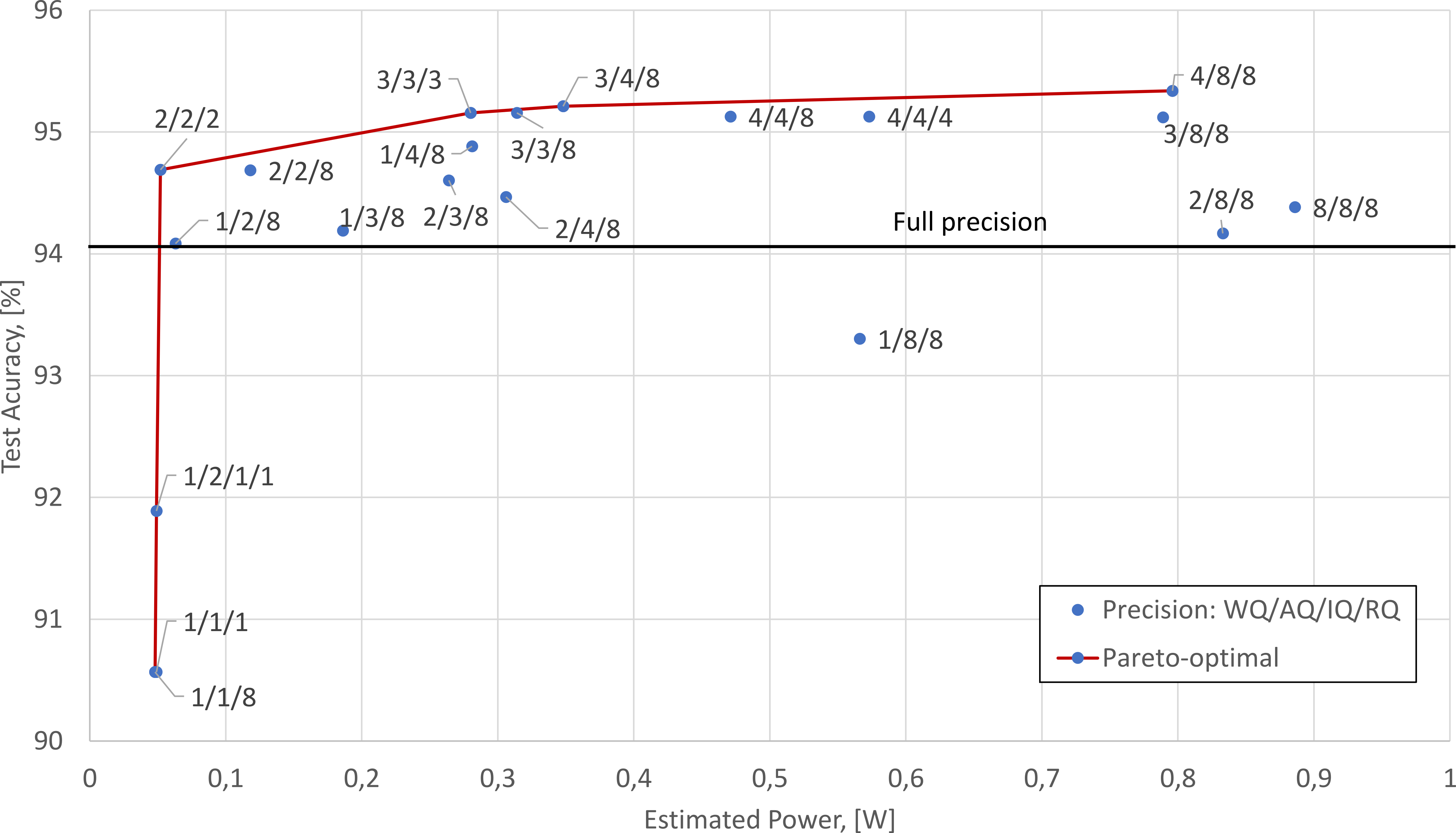}
	\caption{Pareto-frontier for Accuracy against estimated Power consumption}
 	\label{fig:errorPower}
 \end{figure}
 
%The following graphs show the analysis of \gls{ocr} accuracy, achieved during training on the test set, with different precision.

\subsection{Accuracy, hardware cost, and power vs. precision}
The following experiments show the analysis of \gls{ocr} accuracy depending on precision, and corresponding hardware costs and power consumption of a single instance of the accelerator.
Each point represents a combination of number of bits used for representation of weights - WQ, output activations - AQ, input activations - IQ, and recurrent activation - RQ, if different from AQ. In the experiments, the parallelism of the hardware accelerator has been kept constant, with only single \gls{lstm} cell instantiated e.g., \texttt{PE}=1 and full SIMD width e.g., \texttt{SIMD\_RECURRENT}=H, \texttt{SIMD\_INPUT}=I, meaning all inputs and recurrent activations computed in parallel. Target frequency has been set to 200 MHz. All resources are given after post-placement and routing. Fig. \ref{fig:errorBRAM} plots the achieved test accuracy over memory blocks for different precision. Memory blocks are counted as a sum of \#BRAM36 used and 4 times \#URAMs used (URAMs provide exactly 4x the capacity to BRAMs). Fig. \ref{fig:errorLUT} illustrates the achieved test accuracy over the LUT utilization, while Fig. \ref{fig:errorPower} shows test accuracy over the vector-less activity estimated power consumption of the \gls{lstm} accelerator implemented in \gls{pl}. \footnote{The power has been estimated resorting to the Vivado Power Analysis tool on the post-placed netlist, supposing a \textdegree{25}C ambient temperature, 12.5\% of toggle rate and 0.5 of static probability in BRAMs.}

\begin{table*}[t]
\centering
\begin{threeparttable}[b]
\caption{Comparison of implementations for processing dense \gls{lstm} models on FPGA}
\label{tab:comparison_fpga}
\begin{tabular}{llllllllll}
\toprule
                                   & {\cite{chang2015recurrent}}, 2015    & \cite{ferreira2016fpga}, 2016             & \cite{lee2016fpga}, 2016    & \cite{guan2017fpga}, 2017            & \cite{hao2017implementation}, 2017    & \cite{guan2017fp}, 2017          & \cite{rybalkin2017hardware}, 2017    & \cite{han2017ese}, 2017                    & This work                \\
\midrule 
Platform                           & Zynq          & Virtex-7      & Zynq          & Virtex-7      & Zynq          & Stratix-V     & Zynq          & Kintex           & Zynq               \\
                                   & XC7Z020          & XC7VX485T     & XC7Z045          & XCVX485T      & XC7Z020          & GSMD5         & XC7Z045          & XCKU060 & XCZU7EV \\
\midrule 
Model storage                             & off-chip         & on-chip                   & on-chip          & off-chip                 & on-chip                & off-chip               & on-chip          & off-chip                         & on-chip                       \\
Precision\tnote{2}, {[}bits{]}              & 16 fixed         & 18 fixed                  & 4-6 fixed        & 32 float                 & - fixed            & 16 fixed               & 5 fixed          & 12 fixed                         & 1-8 fixed                     \\
Frequency, {[}MHz{]}               & 142              & 141                       & 100              & 150                      & 200              & 150                    & 142              & 200                              & 266                           \\
Throughput\tnote{3}, {[}GOP/S{]} & 0.284            & 4.56                      & -                & 6.87                     & 14.2             & 299                    & 693.12           & 200/1789\tnote{4}                         & 1833    \\
Efficiency\tnote{5}, {[}GOP/J{]}            & 0.15             & -             & -             & 0.37          & -             & 12.63         & 55.88         & 4.88/61.46         & -                \\
\bottomrule
\end{tabular} 
   \begin{tablenotes}
     \item[2] we indicate only precisions that have been demonstrated in the papers. 
     \item[3] throughput has been normalized with respect to 142 MHz, assuming that throughput depends linearly on frequency. 
     \item[4] throughput on a dense model / throughput on a sparse model.
     \item[5] based on only measured power (not estimated).
   \end{tablenotes}
\end{threeparttable}
\end{table*}

%The presented results show that 1-bit weights and 2-bit output activation result in no loss in classification accuracy, and a model with 1/2-bit weights and 3/2-bit activations can even surpass the original full-precision model.
 
\subsection{Throughput scalability vs. precision}

\begin{figure}[]
	\centering
	\includegraphics[width=1.0\columnwidth]{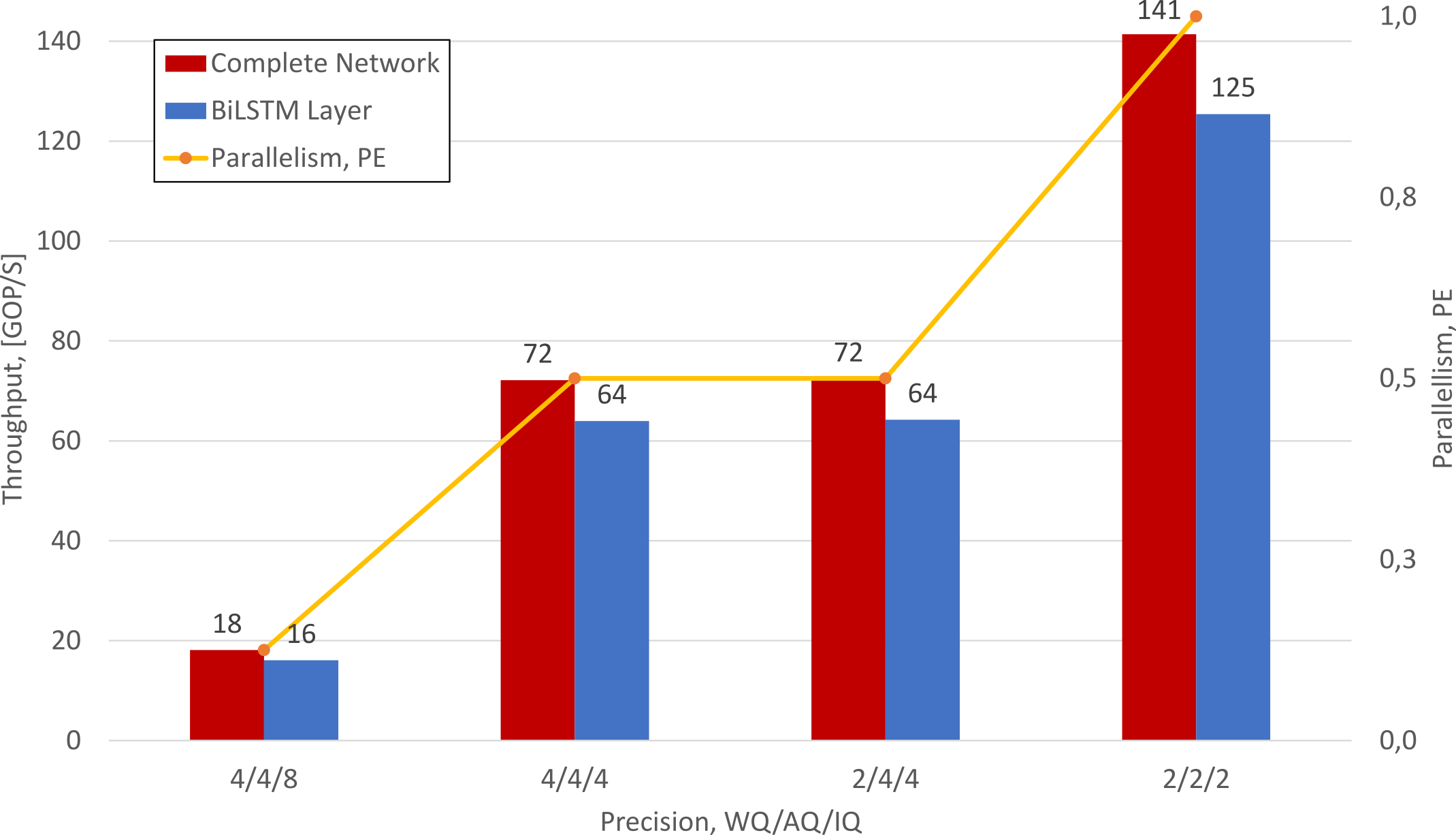}
	\caption{Throughput scalability depending on precision on PYNQ-Z1}
 	\label{fig:Pynq_results}
 \end{figure}

\begin{figure}[]
	\centering
	\includegraphics[width=1.0\columnwidth]{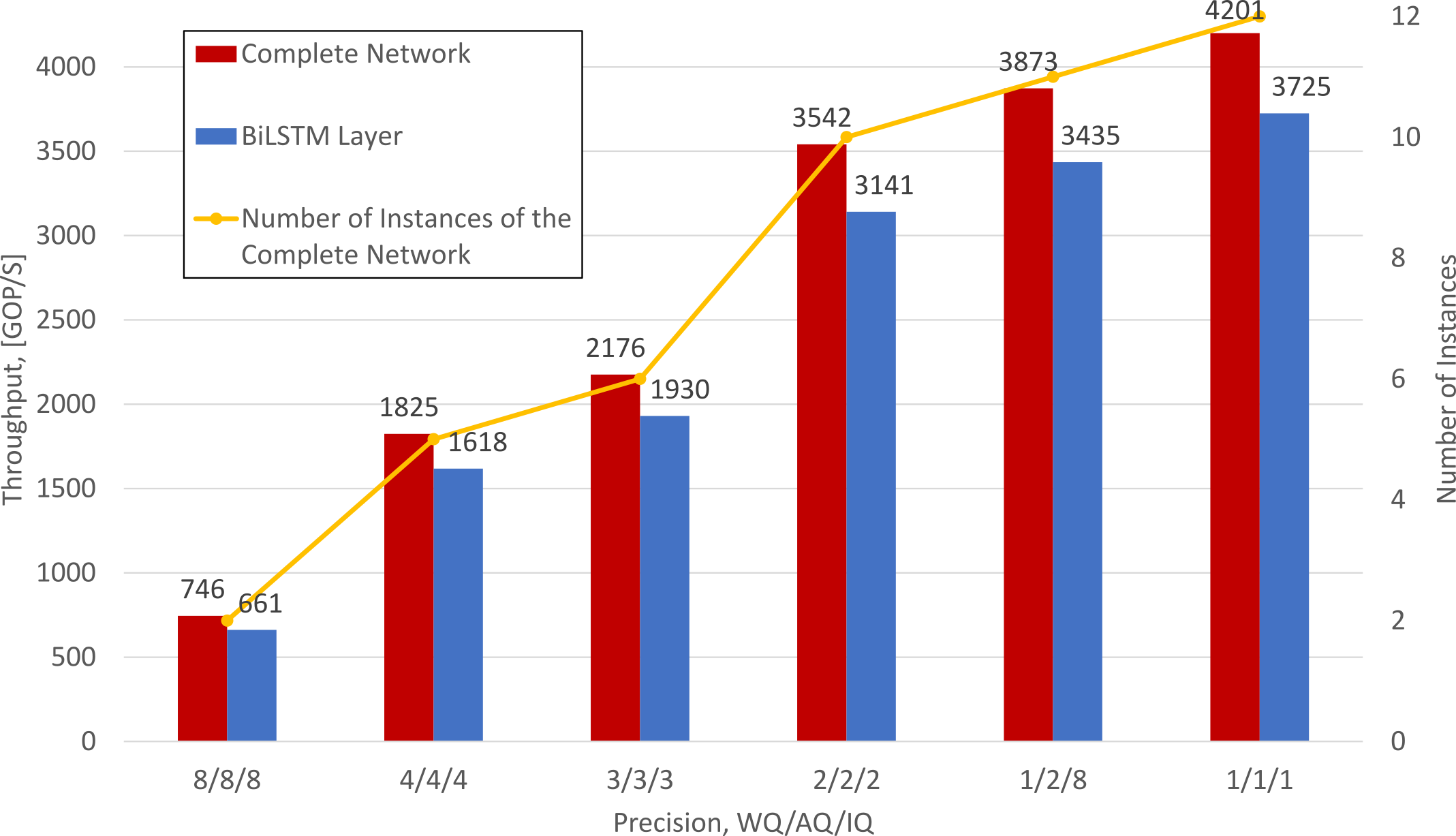}
	\caption{Throughput scalability depending on precision on ZCU104}
 	\label{fig:ZCU104_results}
 \end{figure}

According to Eq. \ref{eq:lstm}, a complexity of a single \gls{lstm} cell is
$2 \times  4 \times (H + I) + 8$, where two comes from multiplication and addition counted as separate operations, four comes from number of gates including input to a memory cell, and eight is a number of point-wise operations. The computed number of operations has to be repeated for each cell and for each timestep of an input sequence. As the proposed architecture implements \gls{bilstm}, the number of operations has to be doubled.
The final formula is: $[2 \times  4 \times (H + I) + 8] \times H \times 2 \times C$. The same applies to the output layer: $[2 \times (2 \times H) + 1] \times K \times C$, where $2 \times H$ gives a number of concatenated outputs from the hidden layers to each unit of the output layer. We neglect a complexity of the rest of the network.
Fig. \ref{fig:Pynq_results} shows a measured on PYNQ-Z1 throughput for different precisions. The inference has been executed on a single image, 520 columns wide, and the accelerator running at 100 MHz \cite{PYNQ-LSTM-REPO}. Due to the limited available resources on PYNQ-Z1, the highest precision solution has been scaled down in parallelism with $F\_s = 8$, while the lowest precision features $PE = 1$, having 8 times parallelism w.r.t. the 4/4/8 configuration.
Fig. \ref{fig:ZCU104_results} shows the estimated throughput on ZCU104 running at 266 MHz, achieved by instead implementing multiple accelerators, where each instance is a complete network depicted in Fig. \ref{fig:inference_topology}. This type of scaling enables parallel execution over multiple images. All configurations have been successfully placed and routed with a realistic set of weights that for some configurations like 2/2/2  has high sparsity. It resulted that some arrays have been optimized out, consequently it allowed to place more instances than it would be possible in the worst case scenario when all arrays are fully implemented. In the case of some configurations like 4/4/4, a maximum number of instances is limited with a number of available BRAMs. In order to implement some arrays using URAM that is not initializable memory, the architecture has been extended with a block that configures arrays at the beginning. That allowed to place more instances by balancing BRAM and URAM usage. 
%This type of scaling will keep a fix latency per each inference image, while enabling parallel execution over multiple images.\\  

%\vspace{-2em}

\subsection{Discussion of the results}

We have shown that our quantized implementation can surpass a single-precision floating-point accuracy for a given dataset when using 1-bit precision for weights and multi-bit quantization for activations. The increase in precision beyond 3 bits is of negligible benefit and only results in increased hardware cost and power consumption. Our experiments confirm observations that have been done in previous works \cite{hou2016loss}, \cite{xu2018alternating} that low bit-width networks can outperform their high-precision counterparts because weight quantization plays a regularization role that prevents model overfitting. 

It has been shown that our proposed approach of quantizing output and recurrent activations differently, namely 1/2/1/1, see Fig. \ref{fig:errorBRAM}-\ref{fig:errorPower}, while being inferior to full-precision, outperforms 1/1/1 configuration by 1.32\% in accuracy with no increase in complexity of the recurrent layer that is responsible for the most of complexity of the network in Fig. \ref{fig:inference_topology}. 

We have demonstrated parameterizable performance scaling for two different parallelization approaches, see Fig. \ref{fig:Pynq_results} and Fig. \ref{fig:ZCU104_results}. The results show flexible customizability of the architecture for different scenarios depending on hardware constrains and required accuracy.

For selected ZCU104 platform and 1/2/8 configuration without accuracy degradation with respect to the full-precision counterpart, the design could achieve a throughput of 3873 GOP/S for the complete network and 3435 GOP/S @ 266 MHz for the \gls{lstm} layer that is the highest with respect to state-of-the-art implementations on FPGAs operating on a dense \gls{lstm} model, see Table \ref{tab:comparison_fpga}. For throughput comparison we use GOP/S that has been used natively in all mentioned papers. Although application level throughput can be a fairer metric, it becomes complicated because of various applications that have been used for benchmarking the designs.

%that is the highest with respect to state-of-the-art implementations of \gls{rnns} on FPGAs operating on a dense model without pruning (see \cref{sec:related_work} for details).

% We have shown that low bit-width \gls{bilstm} can achieve comparable prediction accuracy with full-precision counterparts and even surpass it. In comparison with fully binarized model, increasing bit-width of activations from 1-bit to 2-bit, while still keeping 1-bit weights, leads to significant accuracy increase, approaching the accuracy of model where both weights and activations are 32-bit. However, in the case of \gls{rnns}, even with a single hidden layer, 2-bit recurrent activations result in noticeable increase in hardware complexity. To over overcome the complexity increase, while maintaining the same accuracy as wider bit-width counterparts, we propose to split the recurrent and output activation going to the next layer. By using 1-bit weights, 1-bit recurrent activations and 2-bit output activations, we could reach the same level of accuracy as the multi-bit quantization configurations, while keeping the hardware complexity of the \gls{lstm} layer on the same level with the fully binarized version. 

%We could notice unexpectedly low precision for 2/2/- configuration. That can be explained with unequal representation of a numerical range for negative and positive numbers in a case of 2 bits and power of 2 quantization. It's possible to represent only {-1, -0.5, 0, 0.5}. 

%% file: tex/conclusion.tex
\section{Conclusion}
\label{sec:conclusion}

% We have presented exploration of hardware cost, power consumption and classification accuracy depending on combination of different bit-width for weights, input, in-cell, output and recurrent activations of BiLSTM. We considered binarized and multi-bit quantized BiLSTM model. We do not give specific recommendations on using any of the presented configurations, but provide the results of the experiments as a guideline to a trade-off between accuracy, throughput, and power consumption, meanwhile the choice depends on the application-level requirements.  

This paper presents the first systematic exploration of hardware cost, power consumption, and throughput scalability as a function of precision for \gls{lstm} and \gls{bilstm} in particular. We have conducted an in-depth investigation of precision vs. classification accuracy using a fully hardware-aware training flow, where during training quantization of all aspects of the network are taken into consideration. 

We are providing the first open source HLS library extension of FINN \cite{umuroglu2017finn} for parameterizable hardware architectures of \gls{lstm} layers on FPGAs which offers full precision flexibility and allows for parameterizable performance scaling.% offering different levels of parallelism within the architecture.

Based on this library, we have presented an FPGA-based accelerator for \gls{bilstm} \gls{nn} designed for \gls{ocr} that has achieved the highest throughput with respect to state-of-the-art implementations of \gls{rnns} on FPGAs operating on a dense \gls{lstm} model. We have also demonstrated the first binarized hardware implementation of \gls{lstm} network.